\documentclass[10pt,twocolumn,letterpaper]{article}

\usepackage[pagenumbers]{iccv} 

\usepackage{multirow}
\usepackage{amsmath}

\definecolor{iccvblue}{rgb}{0.21,0.49,0.74}
\usepackage[pagebackref,breaklinks,colorlinks,allcolors=iccvblue]{hyperref}

\def\confName{ICCV}
\def\confYear{2025}

\title{Multimodal Agricultural Agent Architecture (MA3): A New Paradigm for Intelligent Agricultural Decision-Making}

\author{
Zhuoning Xu\textsuperscript{1,2,†}, 
Jian Xu\textsuperscript{1,2,†}, 
Mingqing Zhang\textsuperscript{1,2},
Peijie Wang\textsuperscript{1,2}, 
Chao Deng\textsuperscript{1,2}, 
Cheng-Lin Liu\textsuperscript{1,2,*}\\
\textsuperscript{1}MAIS, Institute of Automation, Chinese Academy of Sciences\\
\textsuperscript{2}School of Artificial Intelligence, University of Chinese Academy of Sciences\\
\textsuperscript{†}Equal contribution, 
\textsuperscript{*}Corresponding author\\
{\tt\small \{xuzhuoning2023,wangpeijie2023,dengchao2023,jian.xu\}@ia.ac.cn}, \\
{\tt\small mingqing.zhang@cripac.ia.ac.cn},
{\tt\small liucl@nlpr.ia.ac.cn}
}

\begin{document}
\twocolumn[{%
\renewcommand\twocolumn[1][]{#1}%
\maketitle
\begin{center}
    \centering
    \captionsetup{type=figure}
    \vskip -0.1in
    \includegraphics[width=\textwidth]{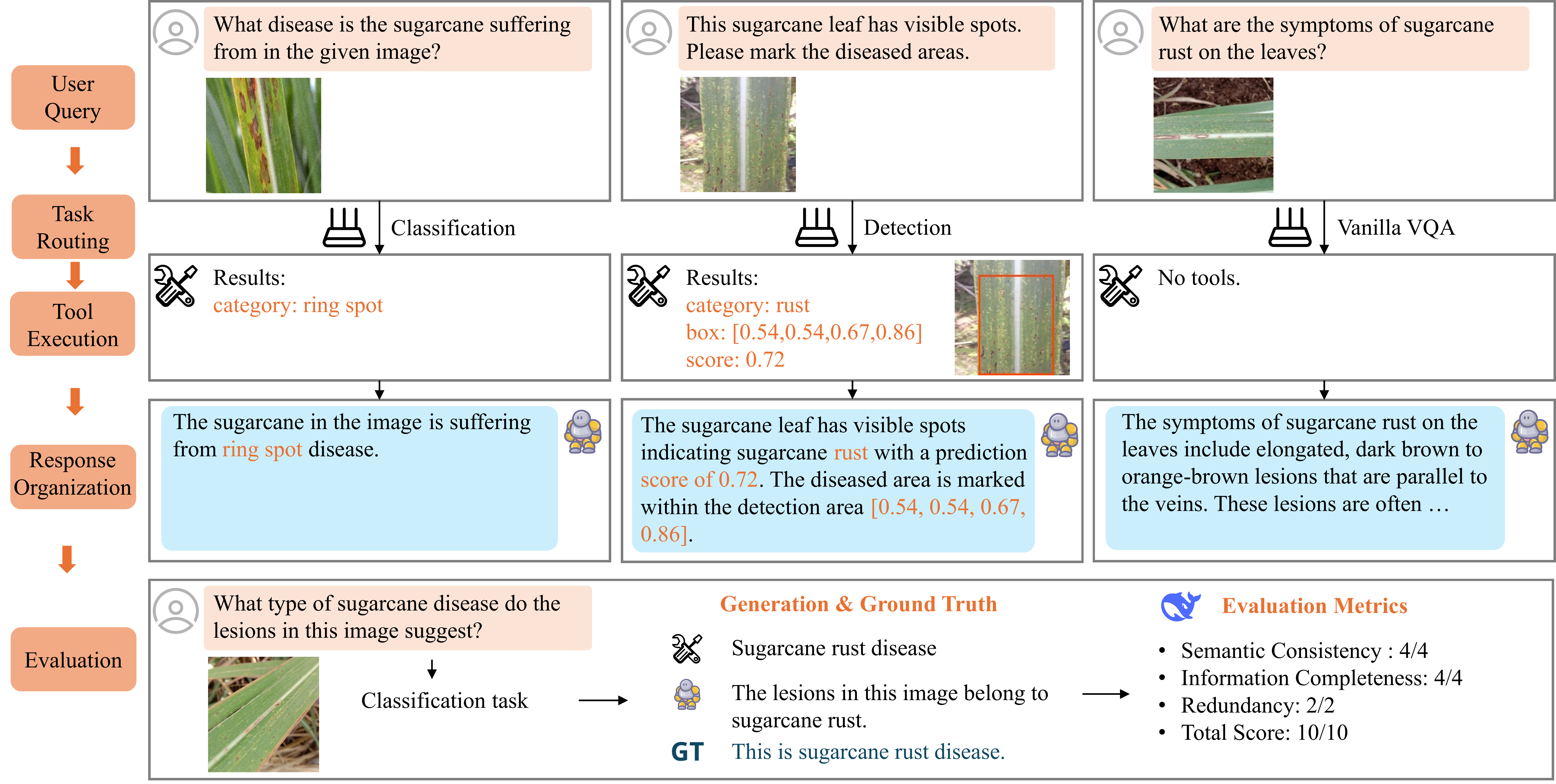}
    \vskip -0.05in
    \captionof{figure}{MA3 is a unified multimodal agricultural agent architecture for intelligent agricultural decision-making, supporting multiple tasks, including disease classification, disease detection, and visual question-answering, through a multi-stage pipeline: (a) task routing, (b) tool execution, (c) response organization and (d) evaluation.
    }
    \label{fig:over}
\end{center}%
}]
\begin{abstract}
\vskip -0.05in
As a strategic pillar industry for human survival and development, modern agriculture faces dual challenges: optimizing production efficiency and achieving sustainable development. Against the backdrop of intensified climate change leading to frequent extreme weather events, the uncertainty risks in agricultural production systems are increasing exponentially. To address these challenges, this study proposes an innovative \textbf{M}ultimodal \textbf{A}gricultural \textbf{A}gent \textbf{A}rchitecture (\textbf{MA3}), which leverages cross-modal information fusion and task collaboration mechanisms to achieve intelligent agricultural decision-making.
This study constructs a multimodal agricultural agent dataset encompassing five major tasks: classification, detection, Visual Question Answering (VQA), tool selection, and agent evaluation. We propose a unified backbone for sugarcane disease classification and detection tools, as well as a sugarcane disease expert model. By integrating an innovative tool selection module, we develop a multimodal agricultural agent capable of effectively performing tasks in classification, detection, and VQA.
Furthermore, we introduce a multi-dimensional quantitative evaluation framework and conduct a comprehensive assessment of the entire architecture over our evaluation dataset, thereby verifying the practicality and robustness of MA3 in agricultural scenarios. This study provides new insights and methodologies for the development of agricultural agents, holding significant theoretical and practical implications. Our source code and dataset will be made publicly available upon acceptance.

\end{abstract}    
\section{Introduction}
\label{sec:intro}
 
Agriculture holds irreplaceable significance in global development. It is the cornerstone for ensuring global food security and meeting the basic survival needs of the population, while also serving as a key industry in driving economic growth, promoting social equity and stability. Moreover, agriculture plays a crucial role in addressing climate change, conserving biodiversity, preserving cultural heritage, and fostering global cooperation. Amidst increasing population and resource-environmental pressures, the sustainable development of agriculture has become a core issue in achieving global sustainable development goals~\cite{xie2023crop, xu2024fertilizer, west2014leverage}.

With the rapid development of artificial intelligence, particularly the emergence of multimodal large language models (MLLMs), agricultural practices have encountered unprecedented opportunities for transformation. MLLMs are capable of integrating and analyzing multiple data modalities—such as images, text, and sensor data—demonstrating significant potential in enhancing decision-making efficiency, optimizing resource management, and improving crop yields. Leveraging the powerful capabilities of MLLMs, relevant practitioners and stakeholders can transition from traditional experience-driven approaches to data-driven intelligent decision-making systems. This paradigm shift not only meets the growing food demands of the global population but also effectively addresses challenges such as climate change, resource scarcity, and environmental degradation.

Among the many crops benefiting from AI technologies, sugarcane holds a central position due to its significant economic and agricultural value. As a primary source of sugar and bioenergy, sugarcane is widely cultivated in tropical and subtropical regions. However, its production faces severe threats from a range of diseases, such as yellow leaf disease, smut, mosaic disease, and rust, which can lead to substantial declines in yield and sugar content~\cite{bhuiyan2021sugarcane, huang2018diagnosis, gaur2009detection, lu2021comparative}. Consequently, sugarcane research has become a crucial topic in the agricultural field and continues to attract significant attention.

By leveraging machine learning methods, it is possible to combine full-sibling progeny genotyping sequencing techniques to predict single nucleotide polymorphisms associated with brown rust resistance in the sugarcane genome~\cite{aono2020machine}. Additionally, lateral flow immunoassay with conjugated labels can simultaneously detect multiple major viruses responsible for sugarcane mosaic disease~\cite{thangavelu2022ultrasensitive}. Smut, one of the most destructive sugarcane diseases globally, can be addressed by artificially inoculating different varieties of the smut fungus Sporisorium scitamineum to screen for disease-resistant cultivars~\cite{rajput2024screening}.

Despite the ongoing research on sugarcane disease recognition and detection, most studies have focused on a limited number of disease categories~\cite{kuppusamy2025enhancing,kavitha2023neural,tanwar2023deep,narmilan2022detection}, leaving other disease types relatively underexplored.
With the rapid development of MLLMs, significant breakthroughs have been achieved in numerous fields~\cite{mllms1,mllms2,mllms3,mllms4,mllms5}. However, many MLLMs still exhibit considerable limitations in performing fine-grained tasks, such as sugarcane disease classification from images. Additionally, these models currently lack capabilities in downstream tasks, such as object detection. To bridge this gap, we propose MA3, a novel framework designed to integrate MLLMs with intelligent agricultural decision-making. By incorporating domain-specific knowledge with advanced AI technologies, MA3 enables precise disease classification, robust object detection, and intelligent decision support for sugarcane cultivation, demonstrating significant potential in multimodal data integration and downstream task execution.

To support MA3, we construct a multimodal agricultural dataset, which is structured into five key tasks: disease classification, disease detection, tool selection, VQA, and agent evaluation. Based on this dataset, we train a sugarcane disease classifier (SDC) and sugarcane disease object detector (SDOD), along with a trainable router for tool selection. Finally, we conduct a multi-dimensional quantitative evaluation (MQE) over our dataset.
Our key contributions can be summarized as follows:

\begin{itemize} 
    \item We propose \textbf{MA3}, an innovative \textbf{M}ultimodal \textbf{A}gricultural \textbf{A}gent \textbf{A}rchitecture, featuring a lightweight tool selector that dynamically coordinates vision models and large language models. This design significantly enhances the system's capability to handle complex agricultural scenarios through efficient task allocation and cross-modal collaboration.
    
    \item To support MA3 development, we curate a comprehensive multimodal agricultural dataset comprising five specialized components: disease classification, disease detection, tool selection, VQA, and agent evaluation. This integrated data foundation enables advanced agricultural disease analysis and facilitates the development of next-generation agricultural AI systems.
    
    \item We introduce a multi-dimensional quantitative evaluation framework and conduct extensive experiments on our evaluation dataset, including ablation studies. Experimental results demonstrate that MA3 outperforms existing models, validating its effectiveness and practical utility in real-world agricultural applications. 
\end{itemize}

\section{Related Work}
\label{sec:related_work}

\subsection{Tool Learning and Selection Mechanism}
In recent years, tool learning has garnered significant attention as a crucial technique for agents to accomplish complex tasks. The essence of tool learning lies in enhancing the capabilities of agents through tool invocation, enabling them to address intricate multimodal tasks. Current research on tool selection primarily revolves around two paradigms: retriever-based tool selection and LLM-based tool selection~\cite{qu2024tool}. The former relies on predefined rules or vector retrieval to match tools, while the latter leverages the semantic understanding capabilities of LLMs to dynamically select tools. 

Recent studies have demonstrated the substantial potential of tool learning in multimodal scenarios. For instance, LLAVA-PLUS~\cite{liu2025llava} coordinates tool invocation through LLM parsing of user instructions, achieving efficient multimodal task processing. Additionally, the three-stage framework proposed by CLOVA~\cite{gao2024clova}—inference, reflection, and learning—further enhances the continuous adaptation of tools. These studies indicate that LLMs, through tool learning, not only augment their ability to solve complex tasks but also expand their application in cross-modal fusion. To systematically evaluate tool invocation capabilities, researchers have constructed various benchmark datasets. T-Eval~\cite{chen2023t} decomposes the tool invocation process into six dimensions, including instruction following, planning, and reasoning. API-Bank~\cite{li2023api} assesses comprehensive invocation abilities through multi-domain API interactions. CARP ~\cite{zhang2024evaluating} focuses on tool-assisted reasoning for computationally intensive tasks. These benchmark datasets provide essential evaluation criteria for tool learning research.

However, existing methods predominantly rely on LLMs as tool selectors~\cite{wang2023mint, gou2023critic, liu2025llava, gao2024clova, 2023toolformer, 2023toolkengpt}, which poses three major challenges: 1) The hallucination issue of LLMs may lead to tool selection failure or conflicts among multiple tasks; 2) The tool selection capability of LLMs is highly dependent on their own model performance and the designed prompting rules; 3) The high computational cost associated with large model parameters (e.g., a single inference of Qwen2.5-7B requires 420 ms) is difficult to meet the real-time requirements of high-demand scenarios such as agriculture. To address these limitations, we propose MA3. Unlike existing LLM-driven methods, MA3 employs a dedicated tool selector (i.e., the Router illustrated in ~\Cref{fig:ar}) that directly models tool invocation logic from annotated data through supervised learning. This approach reduces model parameters, significantly enhances inference speed, and avoids the hallucination issues of LLMs, thereby improving the accuracy of tool selection. MA3 offers an efficient and reliable task routing mechanism for agricultural agents, enabling seamless collaboration between the visual module and language models, and providing a novel solution for complex task handling in agricultural scenarios.

\subsection{Agricultural LLMs}

In recent years, LLMs have made significant advancements in the agricultural domain and have been systematically applied to agricultural knowledge services. These models integrate human expert feedback mechanisms to address domain-specific agricultural challenges. A comprehensive evaluation of popular LLMs in answering agricultural questions has been conducted, with enhancements in generative performance achieved through Retrieval-Augmented Generation and Ensemble Refinement techniques \cite{silva2023gpt}. 

Leveraging a comprehensive database containing over 1.5 million plant science academic articles, further training of Llama 2 with instruction fine-tuning has been shown to improve understanding of plant science-related topics \cite{yang2024pllama, llama2}. 
A multimodal language model approach incorporating Vision-Language Pretraining techniques has been introduced to learn semantic relationships between multimodal information, achieving $94.84\%$ accuracy on a cucumber disease dataset \cite{cao2023cucumber}. 

These works, utilizing domain-specific pretraining, multimodal learning, and knowledge augmentation techniques, have significantly improved LLMs' understanding of specialized agricultural knowledge, multimodal fusion, and information extraction \cite{agricultural2023llm}. The integration of large-scale agricultural data with vision-language pretraining has further enhanced performance in disease identification and knowledge-based question answering, laying a solid foundation for the development and application of AI-driven agricultural systems.

However, existing agricultural LLMs and multimodal agricultural LLMs are primarily designed for solving single tasks. There is currently no unified intelligent agent architecture capable of addressing complex agricultural challenges, nor are there standardized benchmark datasets for evaluation. To bridge this gap, we propose MA3, along with a multi-dimensional quantitative evaluation framework. MA3 enables seamless integration of domain-specific agricultural vision tools and expert models while facilitating the fusion of outputs from multiple models. This architecture enhances the intelligent agent's ability to handle complex tasks while improving model scalability and adaptability. 

\section{Multimodal Agricultural Agent Dataset}
\label{sec:sugarcane_dataset}



\subsection{Sugarcane Disease Image Dataset}

While current vision-language models and multimodal large models exhibit strong performance on general-purpose datasets, they often fail to accurately identify, classify, and detect sugarcane diseases. To address this limitation, we construct a comprehensive sugarcane disease image dataset encompassing both classification and detection tasks, enabling effective model training for domain-specific applications.

Our dataset encompasses 18 distinct sugarcane disease categories, including banded chlorosis, brown spot, brown rust, grassy shoot, pokkah boeng, red rot, sett rot, viral disease, dried leaves, smut, healthy, yellow leaf, rust, ringspot, mosaic, cercospora, bacterial blight, and eyespot. The classification subset comprises approximately 100,000 annotated images, with data augmentation techniques applied to address class imbalance and enhance model robustness.

For the disease detection task, we have more than 60,000 labeled samples, covering the same 18 disease categories. To ensure comprehensive coverage, we implemented a systematic annotation strategy and applied targeted data augmentation methods to underrepresented classes, thereby improving the dataset's diversity and representation of various disease manifestations. The image data statistics are shown in Table~\ref{tab:statistics}.

\subsection{Expert Model VQA Dataset}

To endow our intelligent agent with expert-level knowledge in sugarcane disease diagnosis, this study develops an innovative pipeline for automated generation of domain-specific VQA data. The pipeline architecture comprises four meticulously designed stages:

In the initial phase, we establish a comprehensive expert knowledge base through systematic collection and curation of sugarcane disease-related information from authoritative agricultural resources and scientific literature. This process incorporates rigorous data cleaning protocols and multi-stage validation procedures to ensure the accuracy, reliability, and domain relevance of the compiled knowledge.

Subsequently, we implement a sophisticated data fusion mechanism that strategically integrates image-label categories with their corresponding disease-specific expert knowledge. This fusion process, guided by domain expertise, creates a robust semantic foundation for generating high-quality, contextually relevant VQA pairs.

For the VQA generation phase, we conduct a comprehensive model evaluation comparing several state-of-the-art language models. While both ChatGPT and Qwen2-VL-7B demonstrate comparable performance in generating sugarcane disease VQA data from fused expert knowledge and image labels, we select the Qwen2-VL-7B model for our implementation. This open-source multilingual model, supporting both Chinese and English, is chosen based on its optimal balance between performance and computational efficiency, coupled with the advantages of open-source adaptability for domain-specific fine-tuning.

The final stage incorporates a rigorous secondary data cleaning process, employing both automated and manual verification methods to ensure the quality, accuracy, and domain relevance of the generated VQA pairs. Through this optimized pipeline, we successfully construct a substantial, high-quality VQA dataset, with both Chinese and English subsets exceeding 80,000 samples each.
The specific data statistics are presented in Table~\ref{tab:statistics}.

This dataset not only provides a valuable resource for training and evaluating agricultural vision-language models but also demonstrates the effectiveness of our automated pipeline in generating domain-specific VQA data at scale. The selection of Qwen2-VL-7B, validated through comparative analysis, represents a strategic balance between model performance and practical implementation considerations in agricultural AI applications.

\subsection{Tool Selection Data}

\begin{table}[t]
    \centering
    \resizebox{0.45\textwidth}{!}{
            \begin{tabular}{lr}
            \toprule
            \textbf{Statistic} & \textbf{Number} \\
            
            \midrule
            \textbf{Classification} \\
                ~- train & 86,006 (80.0\%) \\
                ~- val & 10,746 (10.0\%)\\
                ~- test & 10,762 (10.0\%)\\
                ~Total & 107,514 \\
              
            \midrule
            \textbf{Detection} \\
                ~- train & 53,666 (78.0\%) \\
                ~- val & 7,195 (10.5\%)\\
                ~- test & 7,923 (11.5\%)\\
                ~Total & 68,784 \\
                
            \midrule
            \textbf{Tool Selection (Router)} \\
            English & 13,357 \\
                ~- train & 11,416 (85.5\%) \\
                ~- test & 1,941 (14.5\%)  \\
            Chinese & 13,429 \\
                ~- train & 11,449 (85.3\%) \\
                ~- test & 1,980 (14.7\%)  \\

            \midrule
            \textbf{Visual Question-Answering} \\
            English & 85,918\\
            Chinese & 80,556\\

            \bottomrule
        \end{tabular}}
        \captionof{table}{Data statistics based on different tasks and splits.}
        \label{tab:statistics}
        \vspace{-0.4cm}
\end{table}

To facilitate effective processing of downstream tasks, we constructed a bilingual (Chinese and English) prompt text dataset, meticulously partitioned into training and testing sets. Our tool selection framework encompasses three distinct task categories, each designed to optimize the handling of specific query types:

\begin{itemize} 
   \item \textbf{Direct Processing by the Expert Model:} This category is designated for queries requiring domain-specific knowledge interpretation, including: \textit{Disease-specific knowledge inquiries},  \textit{General agricultural knowledge questions}, and \textit{Queries unrelated to classification or detection tasks}.

    \item \textbf{Sugarcane Disease Classification:} Reserved exclusively for queries necessitating disease categorization, this task handles:  \textit{Pure disease classification questions}, \textit{Disease identification requests}, and \textit{Symptom-based classification queries}.  
    
    \item \textbf{Sugarcane Disease Detection:} This comprehensive category processes detection-related queries, including: \textit{Pure disease detection requests}, \textit{Combined classification and detection queries}, and \textit{Localization and identification tasks}. 
\end{itemize}

Through this annotated dataset, agricultural intelligent agents can learn to accurately discern query intent and dynamically invoke the appropriate downstream tools, such as the expert model, classifier, or detector. This capability is crucial for building efficient and precise agricultural AI systems, enabling the deep integration of domain knowledge with task-specific functionalities. The prompt text dataset examples are detailed in the ~\Cref{subsubsec:tool_data_examples}. 



\label{sec:intro}
\begin{figure*}[t]
  \centering
   \includegraphics[width=1.0\linewidth]{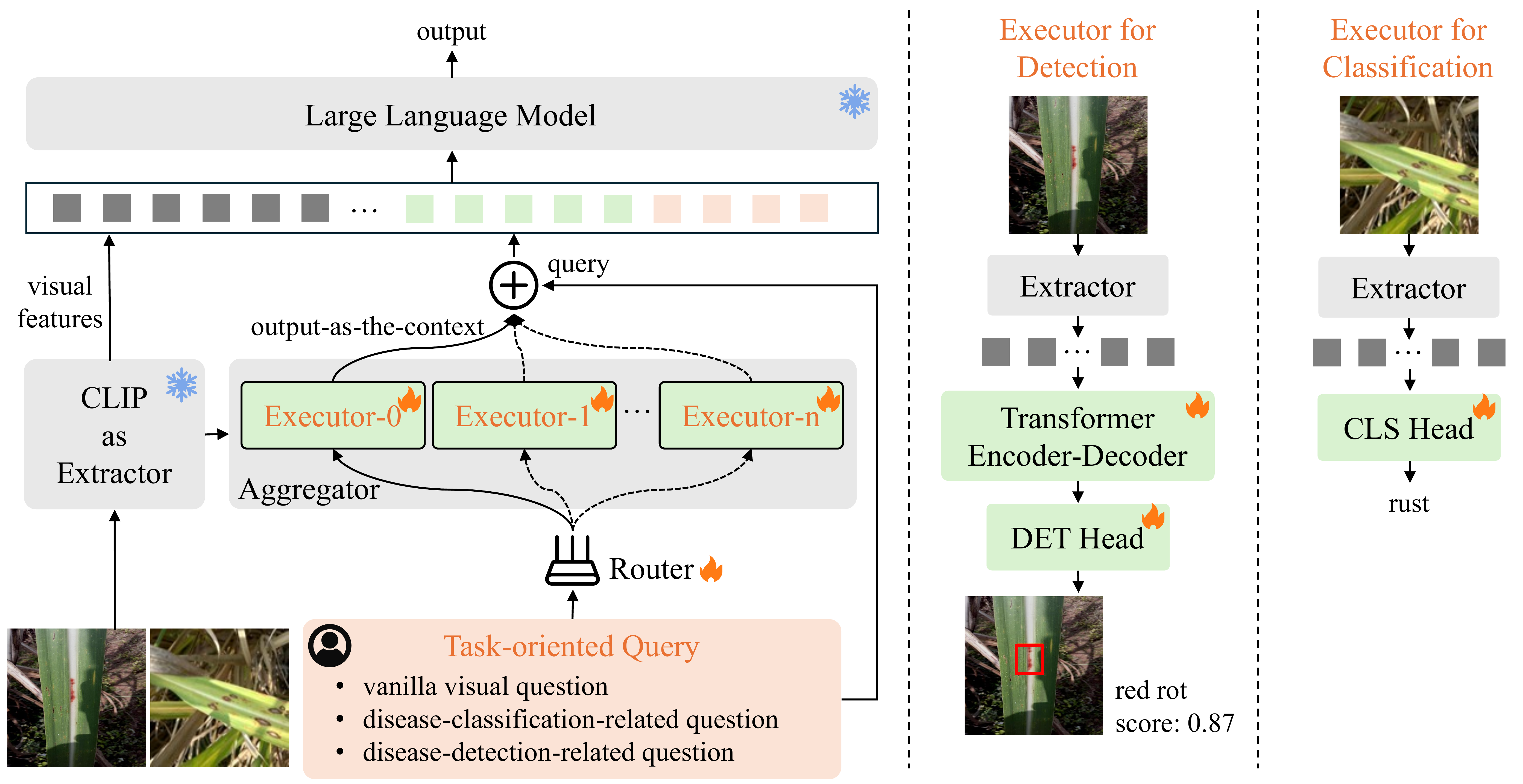}

   \caption{Multimodal Agricultural Agent Architecture (MA3). The MA3 architecture employs a Router to dynamically select among classification tools, detection tools, and the expert model, integrating their outputs with the input text and image before feeding them into the LLM.}
   \label{fig:ar}
   \vspace{-0.4cm}
\end{figure*}

\section{Multimodal Agricultural Agent Architecture (MA3) }





\subsection{System Architecture}

We propose MA3, a comprehensive framework integrating four core components: a router for tool selection, a sugarcane disease classifier, a sugarcane disease detector, and a VQA-fine-tuned expert model. This architecture is designed to enable precise disease analysis and expert-level knowledge dissemination through a modular and extensible pipeline, as shown in Figure~\ref{fig:ar}.

Upon receiving user inputs—comprising both sugarcane disease images and textual queries—the router performs intent classification to determine the optimal processing path. Specifically, it dynamically decides whether to:
1) Directly invoke the fine-tuned expert model for domain-specific knowledge responses;
2) Activate the SDC for symptom-based categorization;
3) Engage the disease detector for lesion localization and identification.

Concurrently, the input images are encoded through a vision encoder, and the resulting visual representations are routed according to the router's classification output. The system then fuses the outputs from the respective modules with the original textual input, enabling context-aware disease analysis that combines visual evidence with semantic understanding. This architecture endows MA3 with three key capabilities:

\begin{itemize} 
   \item \textbf{Accurate Disease Classification:} Precise identification of sugarcane diseases based on visual symptoms and contextual cues.

    \item \textbf{Targeted Lesion Detection:} Localization and characterization of disease-specific patterns in images, even under challenging conditions.  
    
    \item \textbf{Expert Knowledge Integration:} Provision of domain-specific insights through the fine-tuned multimodal model, ensuring reliable and actionable recommendations. 
\end{itemize}

The modular design of MA3 not only ensures flexibility in handling diverse query types but also supports seamless extension to additional crops or disease categories. The Router, as the central decision-making component, plays a critical role in orchestrating the tool selection process, enabling efficient resource allocation and task-specific optimization. This design philosophy makes MA3 a scalable and adaptable solution for real-world agricultural applications.

\subsection{Basic Visual Tool Structure}
\subsubsection{Backbone}

CLIP~\cite{radford2021learning} is a vision-language model trained on a dataset comprising 400 million image-text pairs using contrastive learning. It demonstrates exceptional performance in zero-shot text-image retrieval, zero-shot image classification, and open-domain detection and segmentation tasks. Given its robust capabilities, we adopt CLIP-ViT as the shared backbone for both the SDC and the SDOD.

\subsubsection{Sugarcane Disease Classifier}
CLIP-ViT is pre-trained on a large image dataset and provides a powerful visual representation suitable for tasks such as sugarcane disease classification. The classifier is built by combining CLIP-ViT with a linear classification layer to achieve accurate disease detection. The detailed architecture is illustrated in Figure~\ref{fig:ar}. 

The classification loss function is defined as follows,
\begin{align*}
w_y = \min\left(\sqrt{\frac{N}{N_y}}, 10\right)
\end{align*}
\begin{align*}
\mathcal{L}(\mathbf{p}, y) = -w_y \cdot \log\left(\frac{\exp(p_y)}{\sum_{j=1}^C \exp(p_j)}\right)
\end{align*}
where $C$ is the number of classes, 
\( \mathbf{p} = [p_1, p_2, \dots, p_C] \) is the predicted probability distribution, \( y \in \{1, 2, \dots, C\} \) is the ground truth label, and $w_y$ is the class weight, with N being the total number of samples and $N_y$ the number of samples for $class_y$.

\subsubsection{Sugarcane Disease Object Detector}
We extend the DETR~\cite{DETR} architecture and replace its image encoder with CLIP-ViT. The extracted features interact with the object query through a Transformer decoder to produce detection results, including category labels and bounding boxes. In this setting, the "no object" class corresponds to the image background because the dataset is fully annotated. We refer to the DETR model architecture, hence we also utilize the Hungarian loss~\cite{DBLP:conf/cvpr/StewartAN16} for our computations. The detailed architecture is illustrated in Figure~\ref{fig:ar}.

\subsection{Intelligent Agent Brain}
\label{subsec:agent_brain }

\subsubsection{Router}
\label{subsubsec:tool_selector }

Driven by user prompts, the tool selection task is fundamentally a text classification problem aimed at accurately mapping queries to downstream processing modules. Given the high density of domain-specific terminology and semantic complexity in agricultural contexts, we employ  BERT~\cite{bert} as the core classifier. As a bidirectional pre-trained language model based on the Transformer architecture~\cite{transformer}, BERT effectively captures long-range semantic dependencies prevalent in agricultural consultations by simultaneously parsing the left and right contextual information of words. Its pre-training and fine-tuning paradigm enables language representations pre-trained on large-scale general corpora to quickly adapt to fine-grained classification requirements in agriculture, while its bidirectional attention mechanism accurately identifies core intents in complex queries.

Our classifier dynamically routes user queries to three processing paths:

\textbf{Expert Model Response:} Handles knowledge-based inquiries (e.g., disease pathology mechanisms).

\textbf{Visual Classifier Invocation:} Triggers disease recognition based on symptom images.

\textbf{Object Detector Activation:} Executes lesion localization or hybrid tasks.

Experimental validation (Section \ref{sec:experiment_tool}) shows that BERT-base outperforms Qwen2.5-7B in terms of accuracy on the validation set, with an inference speed that is 130 times faster. This design optimizes the trade-off between accuracy and efficiency, demonstrating the feasibility of lightweight pre-trained models for agricultural tool scheduling.
By integrating domain knowledge with efficient tool routing, the framework enables real-time agricultural decision-making systems, even in resource-constrained environments. Experimental results confirm its high accuracy and significantly reduced deployment costs, paving the way for practical adoption in the agricultural sector.

\subsubsection{Output Fusion}

To effectively integrate the tool outputs with the input text, we have chosen to use MLLM or LLM for the fusion process. We found that the fine-tuned expert model was not able to execute the fusion instructions effectively. Therefore, we selected the open-source model Qwen2.5-32B, which has strong instruction-following capabilities, to perform the fusion of inputs and outputs. The fusion process includes:  
1) Alignment: Structuring the tool outputs into a unified format compatible with the fusion model;  
2) Context Integration: Incorporating the input queries to maintain task relevance;  
3) Knowledge Synthesis: Generating a final response that incorporates tool information and is contextually appropriate.

\section{Experiment}


\subsection{Sugarcane Disease Classifier}
Adopting a transfer learning strategy, we froze the vision tower of CLIP and fine-tuned only the classification head. The model was trained on two A6000 GPUs, with weights saved at peak validation accuracy. Final evaluation on the test set demonstrated that the pretrained visual features effectively adapted to agricultural disease recognition tasks, achieving a precision of 96.2\% on the test set.

\subsection{Sugarcane Disease Object Detector}
Building on the frozen vision tower of CLIP, we fine-tuned the transformer layers and the feed-forward network (FFN) using the AdamW optimizer~\cite{AdamW}, with a learning rate set to $10^{-4}$. The detector was trained for on dual A6000 GPUs and evaluated using COCO API protocols. Given the inherent challenges of agricultural imagery, including inconsistent annotation quality and variable lesion sizes, we emphasize mAP@0.4 as a critical metric for practical field applications, alongside the standard mAP@0.5. The experimental results are summarized in Table~\ref{tab:det}.

\subsection{Router}
\label{sec:experiment_tool}

We train and test the BERT model using our Chinese and English prompt datasets separately, and compared its performance with two LLM variants: Qwen2.5-7B~\cite{qwen2.5} and Qwen2.5-32B~\cite{qwen2.5}. Through iterative optimization of prompt rules and model scaling, we enhanced the tool selection capabilities of the LLMs. The experimental results are shown in Table~\ref{tab:bert}.

The results demonstrate that our fine-tuned BERT model outperforms both LLM variants on Chinese tasks, achieving a classification accuracy of 99.34\%, which is superior to both Qwen2.5-32B and Qwen2.5-7B. For English tasks, BERT achieves an accuracy of 99.12\%, which is comparable to Qwen2.5-32B and 13.5\% higher than Qwen2.5-7B.

BERT's exceptional accuracy, coupled with its significantly smaller model size, makes it more suitable for tool selection tasks than LLMs. Specifically, BERT’s lightweight architecture reduces both training and inference time costs while maintaining high performance. In contrast, the effectiveness of LLMs as tool selectors heavily relies on the language model’s capabilities and the design of prompt rules. As the number of downstream tasks increases, designing clear and unambiguous prompt rules becomes challenging due to potential keyword overlaps between tasks, thereby increasing the complexity of rule design. 

This experiment successfully demonstrated the efficacy of lightweight pre-trained language models in the domain of agricultural tool scheduling by employing the BERT model. The BERT model, with its superior ability to accurately understand user intents, enabled efficient collaborative optimization between the agricultural expert system and the visual module. This achievement provides an efficient and viable framework for the deployment of real-time agricultural decision-making systems.

\begin{table}[t]
  \centering
  \setlength{\tabcolsep}{2.9pt} 
  \begin{tabular}{@{}lcccccccc@{}}
    \toprule
    Model & mAP &\( AP_{\text{40}}\) & \( AP_{\text{50}}\) & \( AP_{\text{75}}\)  & \( AP_{\text{S}}\) & \( AP_{\text{M}}\) & \( AP_{\text{L}}\)\\
    \midrule
    SDOD & 0.325 & 0.718 & 0.643 & 0.287  & 0.031 & 0.190 & 0.490 \\
    \bottomrule
  \end{tabular}
  \caption{SDOD evaluation metrics.}
  \label{tab:det}
  
\end{table}

\begin{table}[t]
  \centering
  \begin{tabular}{@{}lccc@{}}
    \toprule
    Language & Qwen2.5-7B & Qwen2.5-32B & BERT\\
    \midrule
    Chinese & $93.64\%$ & $94.90\%$ & $99.34\%$\\
    English & $87.33\%$ & $97.94\%$ & $99.12\%$\\
    \bottomrule
  \end{tabular}
  \caption{Experimental results of prompt text data on different models.  The optimal performance of BERT.}
  \label{tab:bert}
  \vspace{-0.2cm}
  
\end{table}

\subsection{Multi-Dimensional Quantitative Evaluation}
We leverage DeepSeek-V3~\cite{deepseekv3} as the judge model for automated evaluation across four dimensions: semantic consistency, information completeness, information leakage, and redundancy. The evaluation example is shown in Figure~\ref{fig:ds_evel}.

\begin{table*}[t]
  \centering
  \begin{tabular}{@{}ll|ccccc|cccc|cc@{}}
    \toprule
    \multirow{2}{*}{\textbf{Model}} & \multirow{2}{*}{\textbf{Tools}} & \multicolumn{5}{c|}{\textbf{Classification}} & \multicolumn{4}{c|}{\textbf{Detection}} & \multirow{2}{*}{$\textbf{S}_\textbf{cls}$} & \multirow{2}{*}{$\textbf{S}_\textbf{det}$} \\
    
     & & $Acc$ & $C_{amc}$ & $C_{sc}$ & $C_{ic}$ & $C_{nr}$ & $D_{sc}$ & $D_{ic}$ & $D_{nr}$ & $D_{il}$ & & \\
    
    \midrule
    Llava1.5-13B & w/o & 0.34 & 0 & 0.60 & 0.74 & 0.51 & 0.37 & 0.27 & \textbf{0.83} & \textbf{1} & 0.64 & 0.51 \\
    Llava1.5-13B & w/ & 0.85 & 0.22 & 0.86 & 0.81 & 0.69 & 0.74 & 0.80 & 0.34 & 0.99 & 0.80 & 0.77 \\
    Qwen2.5-32B & w/ & \textbf{0.87} & \textbf{0.87} & \textbf{0.88} & \textbf{0.85} & \textbf{0.79} & \textbf{0.87} & \textbf{0.92} & 0.72 & 0.84 & \textbf{0.85} & \textbf{0.86} \\
    \bottomrule
  \end{tabular}
  \caption{Evaluation results of visual tools based on DeepSeek-V3. Classification task metrics include: $Acc$, \(C_{amc}\), \(C_{sc}\), \(C_{ic}\), and \(C_{nr}\). Detection task metrics include: \(D_{sc}\), \(D_{ic}\), \(D_{nr}\), and \(D_{il}\). \(S_{cls}\) and  \(S_{det}\) represent the comprehensive scores for classification and detection, respectively. abbreviations: amc, anti-misleading capability; ic, information completeness; sc, semantic consistency; nr, non-redundancy; il, information leakage.}
  \label{tab:DSevel}
  \vspace{-0.4cm}
\end{table*}

\subsubsection{Classification Pipeline Evaluation}

We evaluate the classification task over a test set containing 500 images spanning 18 sugarcane disease categories, paired with 500 disease classification prompts that include category-specific descriptions. To ensure data quality, all image-text pairs are manually annotated and validated by domain experts. To assess the model’s robustness against adversarial inputs, we introduce 54 adversarial samples with semantic distractions. Evaluation Metrics:
\begin{itemize}
    \item Accuracy (\textit{Acc}) is one of our classification evaluation metrics, and its calculation formula is as follows: 
    \begin{align*}
    Acc = P_1 \times P_2,
    \end{align*}%
 where \( P_1 \) is the proportion of disease classification texts selected by the tool selector relative to the total number of texts, \( P_2 \) is the accuracy of the model's output in matching the true category of the image.
 
    \item Anti-Misleading Capability (\( C_{amc} \)): Ratio of correct predictions on adversarial samples.

    \item Semantic Consistency (\( C_{sc} \)): Alignment between model outputs and expert-validated reference answers.
    
    \item Information Completeness (\( C_{ic} \)): Coverage of disease categories and diagnostic evidence in the outputs.
    
    \item None-Redundancy (\( C_{nr} \)): Calculated as 1 minus the proportion of irrelevant information in the outputs.
\end{itemize}

\subsubsection{Detection Pipeline Evaluation}

The construction of the evaluation set for the detection pipeline follows a similar approach to that of the classification pipeline. The detection evaluation dataset consists of 200 test samples, each comprising a disease detection prompt, a sugarcane image, and annotated disease regions with corresponding labels. Evaluation Metrics:

\begin{itemize}
 
    \item Semantic Consistency (\( D_{sc} \)) \& Non-Redundancy (\( D_{nr} \)): Same as in classification tasks.
    
    \item Information Completeness (\( D_{ic} \)): Inclusion of detection category, localization regions, and confidence scores.
    
    \item Information Leakage (\( D_{il} \)): Risk of exposing intermediate prediction variables (e.g., detection tool parameters).
\end{itemize}


\subsubsection{Formulation of Evaluation Functions}

\textbf{Classification Task:} 
\begin{align*}
S^i_{\text{cls}} = 0.4\cdot C^i_{\text{sc}} + 0.4\cdot C^i_{\text{ic}} + 0.2\cdot C^i_{\text{nr}}
\end{align*}%
where \( C^i_{\text{sc}} \) denotes the text-image alignment score for sample i, \( C^i_{\text{ic}} \) represents the coverage of key pathological features for sample i, \(C^i_{\text{nr}}\) is the penalty term for irrelevant statements in sample i.

\noindent \textbf{Detection Task:} 
\begin{align*}
S^i_{\text{det}} = 0.4\cdot D^i_{\text{sc}} + 0.3\cdot D^i_{\text{ic}} + 0.2\cdot D^i_{\text{il}} + 0.1\cdot D^i_{\text{nr}}
\end{align*}%
where \( D^i_{\text{il}}\) indicates the risk coefficient of information exposure for sample 
i.

$S_{cls}$ reflects the dual requirements of precision and conciseness for the agent in agricultural classification tasks, while 
$S_{det}$ not only captures these aspects for detection tasks but also embodies the balance between security and efficiency in the agent system. The proportions within both $S_{cls}$ and $S_{det}$ are parameterized based on the contribution degrees to agricultural intelligent decision-making, reflecting the model's performance across different decision-making dimensions.

To evaluate the impact of tool integration on visual tasks, we employ the widely-used MLLM, LLaVA-1.5-13B~\cite{liu2024llava1.5}, as a baseline comparison. Since the Qwen series demonstrates superior performance in VQA data construction and tool selection tasks, we select it as our primary LLM. Experimental results, as shown in Table~\ref{tab:DSevel}, reveal that LLaVA-1.5-13B with tools significantly outperforms the baseline (without tools) in three key metrics: $Acc$, $S_{cls}$, and $S_{det}$. This improvement is attributed to the fact that the baseline LLaVA-1.5-13B, without fine-tuning or training, fails to accurately recognize all sugarcane disease categories. Notably, the baseline achieves a perfect information leakage score ($D_{il}$=1.0) due to its isolation from tool-related information, ensuring complete information security at the cost of task-solving capability.

In contrast, Qwen2.5-32B outperforms both LLaVA variants in four critical metrics: $Acc$, $C_{amc}$, $S_{cls}$, and $S_{det}$. This advantage stems from its stronger language understanding capabilities, which enable more comprehensive utilization of prior information from visual tools, thereby enhancing its robustness against interference. However, this deep integration also increases the risk of information leakage, as reflected in its $D_{nr}$ value of 0.72, compared to 0.99 for LLaVA with tools. This highlights the inherent trade-off between performance and security in tool-augmented systems.

\section{Conclusion}

In this study, we propose MA3, a novel paradigm for intelligent agricultural decision-making designed to address key challenges in agricultural knowledge question answering (QA) and visual analysis. We construct a multimodal agricultural dataset aligned with the framework, covering five core tasks: classification, detection, tool selection, visual question answering (VQA), and agent evaluation. MA3 integrates domain-specific vision tools with an expert model equipped with agricultural disease knowledge through a lightweight router, achieving robust performance in VQA, disease classification, and detection tasks.

Our key innovation lies in replacing traditional large language model (LLM)-based tool selection with a supervised BERT model, achieving both high accuracy and efficient inference. This design not only overcomes the limitations of large language models (e.g., hallucination issues and high computational costs) but also provides a scalable architecture for future extensions such as semantic segmentation and image generation. However, the current implementation relies on supervised data, which may limit its adaptability to unseen domains. Future work will explore semi-supervised learning to enhance generalization capabilities. Additionally, we plan to optimize the Router for enhanced multi-task collaboration, improving effectiveness in diverse agricultural scenarios.

{
    \small
    \bibliographystyle{ieeenat_fullname}
    \bibliography{main}
}
\clearpage
\appendix
\setcounter{page}{1}
\maketitlesupplementary

\setcounter{section}{0}
\renewcommand{\thesection}{\Alph{section}}  

\section{Data}
\subsection{Image Data and Analysis}
\subsubsection{Image Data Examples}
\label{subsubsec:image_data_examples}

\begin{figure*}[t]
  \centering
   \includegraphics[width=0.93\linewidth]{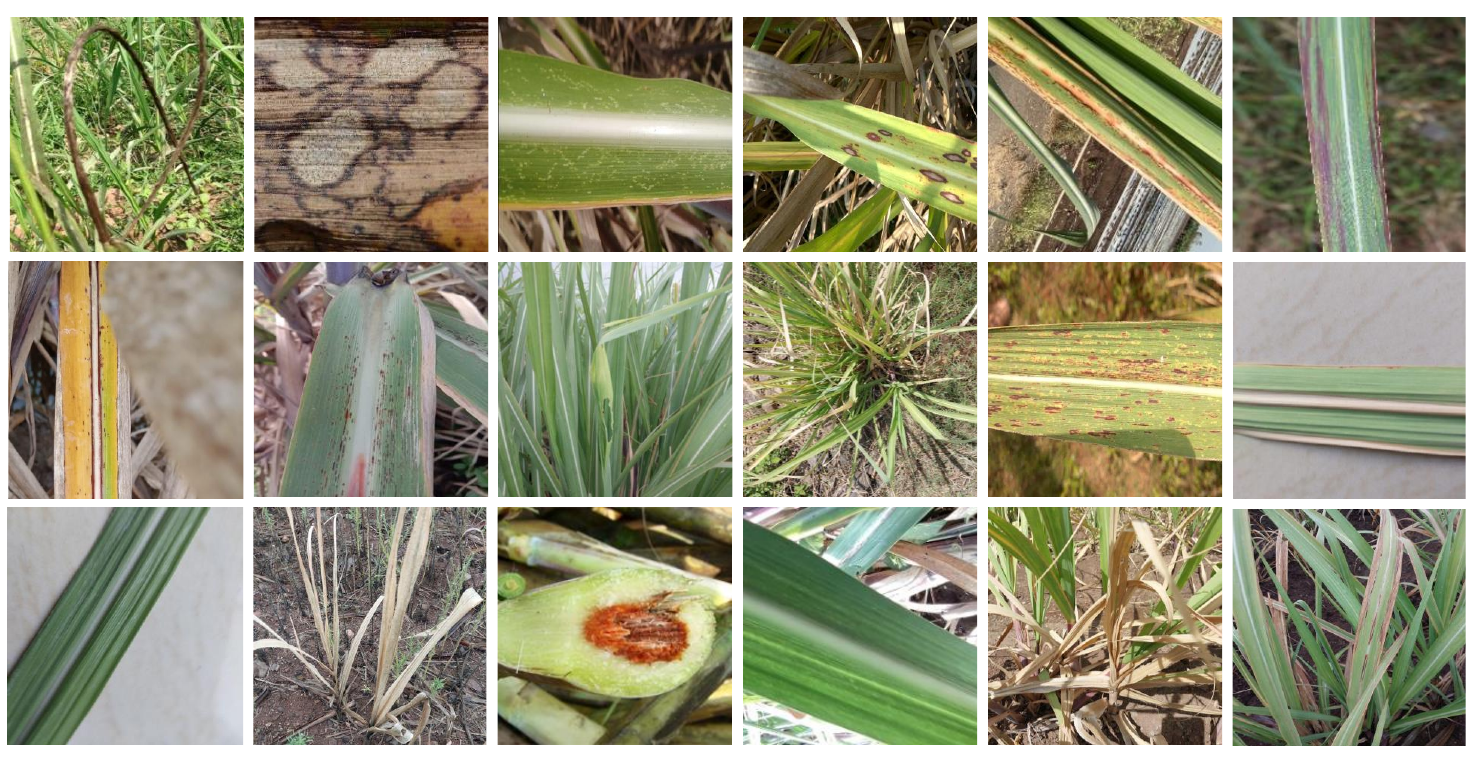}

   \caption{Examples of images of 18 sugarcane diseases. }
   \label{fig:image}
\end{figure*}

We initially collected 30,000 annotated samples for object detection, covering 9 sugarcane disease categories. To address data scarcity, we manually annotated the remaining 9 disease categories and performed data augmentation for underrepresented classes. This process resulted in a final dataset of over 60,000 annotated samples spanning all 18 disease categories. All image data were uniformly resized to 336×336 pixels before being input to the vision tools. The image data is illustrated in Figure~\ref{fig:image}. 

\subsubsection{Data Analysis}
\begin{figure*}[t]
  \centering
  \begin{subfigure}{0.48\linewidth}
    \includegraphics[width=1\linewidth]{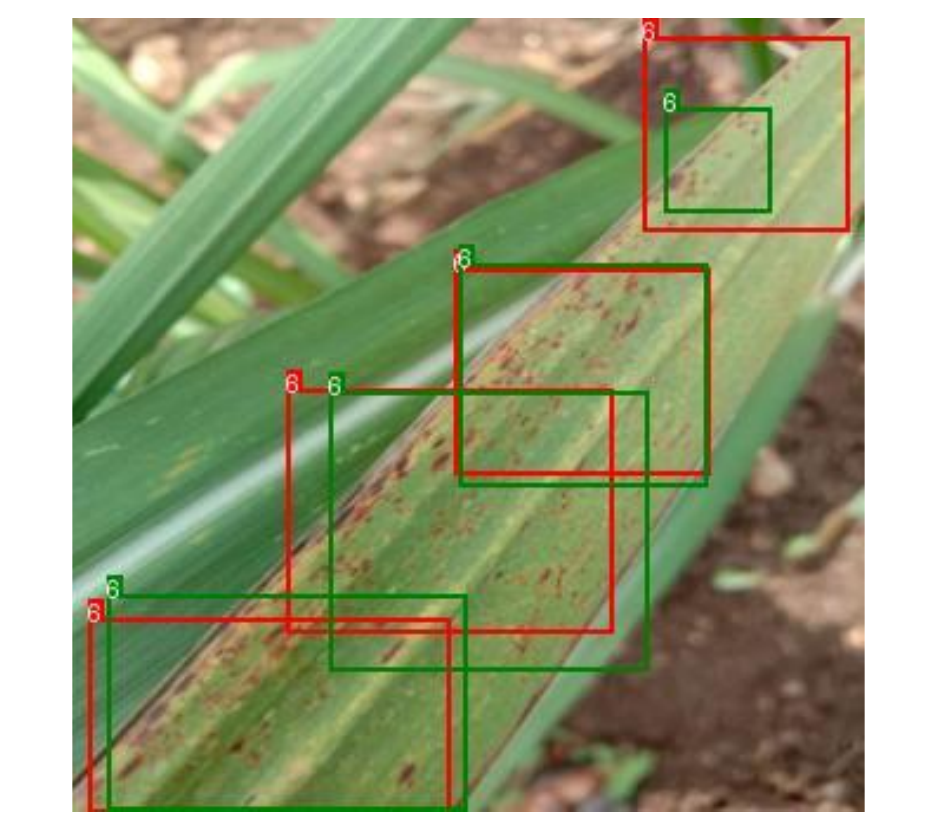}
    \caption{Diseased sugarcane leaves. }
    \label{fig:short-a}
  \end{subfigure}
  \hfill
  \begin{subfigure}{0.48\linewidth}
    \includegraphics[width=1\linewidth]{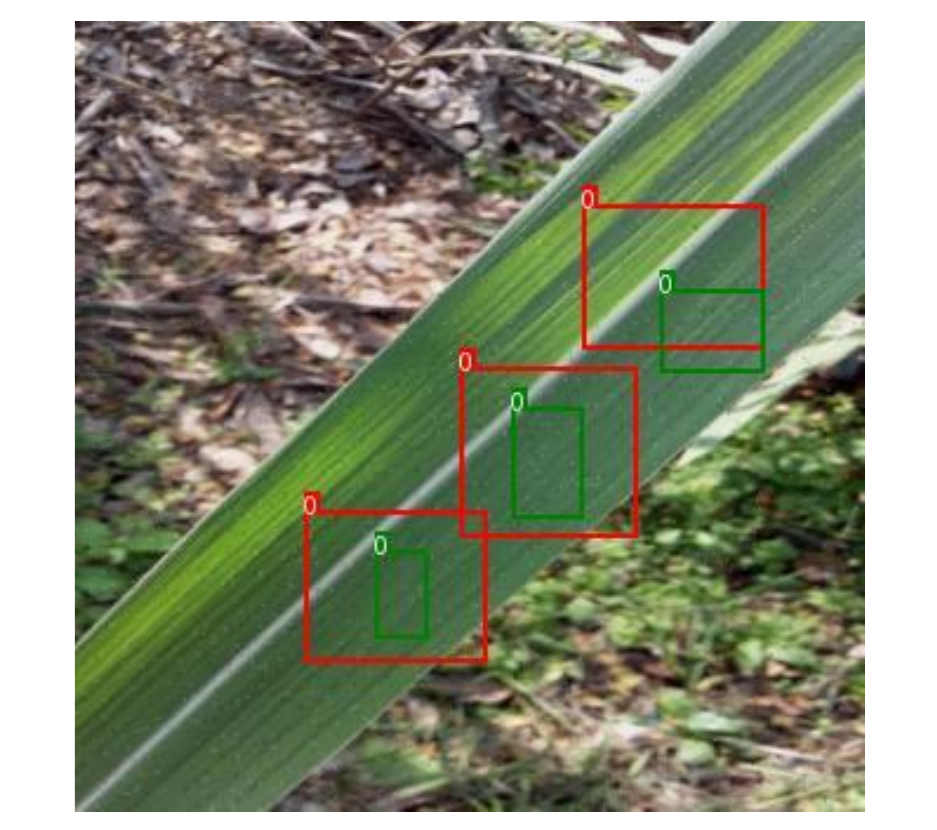}
    \caption{Healthy sugarcane leaves. }
    \label{fig:short-b}
  \end{subfigure}
  \caption{Prediction results for healthy sugarcane and diseased sugarcane.}
  \label{fig:short}
\end{figure*}

In our detection dataset, a single sugarcane leaf may contain multiple bounding box annotations. Therefore, during model prediction, any predicted box that contains a candidate region of sugarcane disease or healthy tissue is considered correct from a human evaluator's perspective, regardless of its size or deviation from the ground truth box. However, from a model evaluation perspective, we aim for predicted boxes to closely match the ground truth boxes, as this indicates better model performance. To balance human assessment and model performance evaluation, we appropriately lower the IoU threshold during model evaluation. As shown in Figures~\ref{fig:short}, the Intersection over Union (IoU) value between the predicted box and the ground-truth box in the upper right corner of the diseased sugarcane leaf does not reach 0.5. However, since the predicted box encompasses the region of disease, the prediction is considered correct. For healthy sugarcane leaves, a prediction is deemed correct from the perspective of human evaluation as long as the predicted region is located on the leaf, even if there is a deviation from the ground-truth area.





\subsection{Expert Model VQA Dataset Construction}
The VQA dataset constructed from sugarcane disease image data serves two primary purposes:
\begin{itemize} 
   \item \textbf{Category Alignment:}  Ensuring accurate mapping between visual disease symptoms and their corresponding categories.

    \item \textbf{Expert Knowledge Alignment:} Incorporating domain-specific contextual information related to sugarcane diseases.  
    
\end{itemize}

During dataset construction, we integrate images and their corresponding labels with our disease knowledge base, enabling the MLLM to generate contextually relevant data guided by prior knowledge. However, the data generation process reveals two key challenges: 
hallucination issues and format inconsistencies

To address these issues, we implement a two-stage data cleaning pipeline: 1)Content Filtering: Removing hallucinated or irrelevant outputs based on domain-specific rules. 2)Format Standardization: Enforcing consistent output structures to ensure data completeness and usability.

\begin{figure*}[h]
  \centering
   \includegraphics[width=1.0\linewidth]{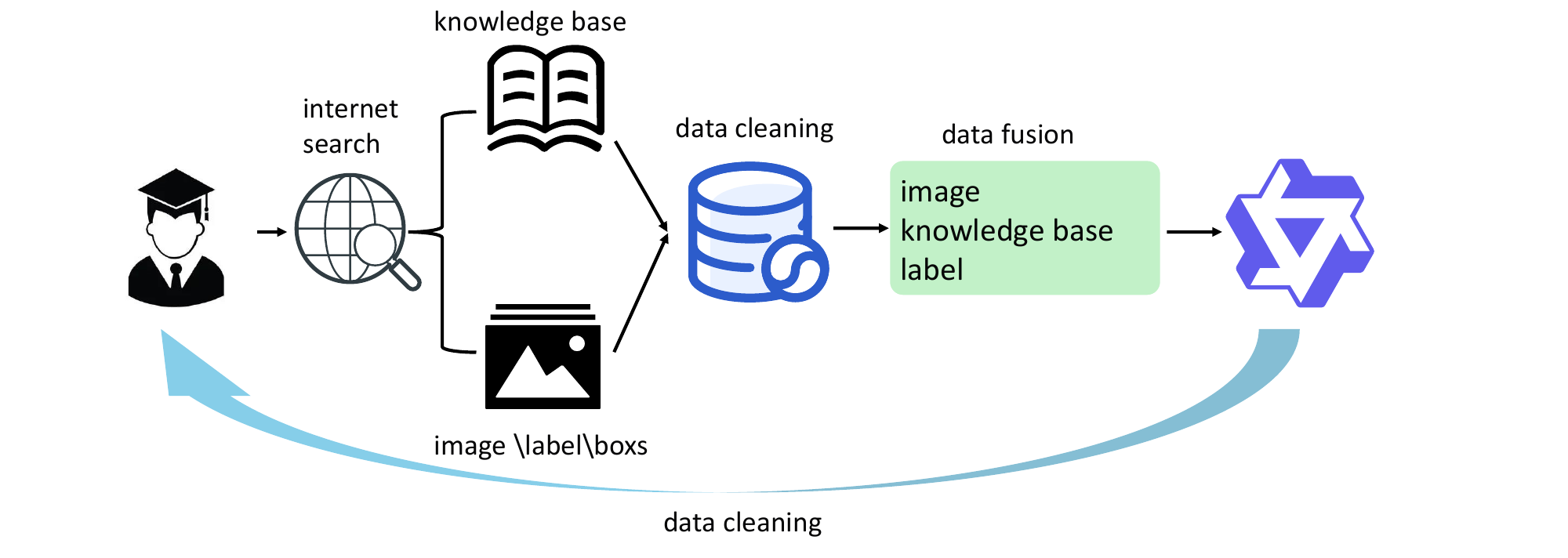}

   \caption{VQA data construction pipeline. }
   \label{fig:data_con}
\end{figure*}

\begin{figure*}[h]
  \centering
   \includegraphics[width=1\linewidth]{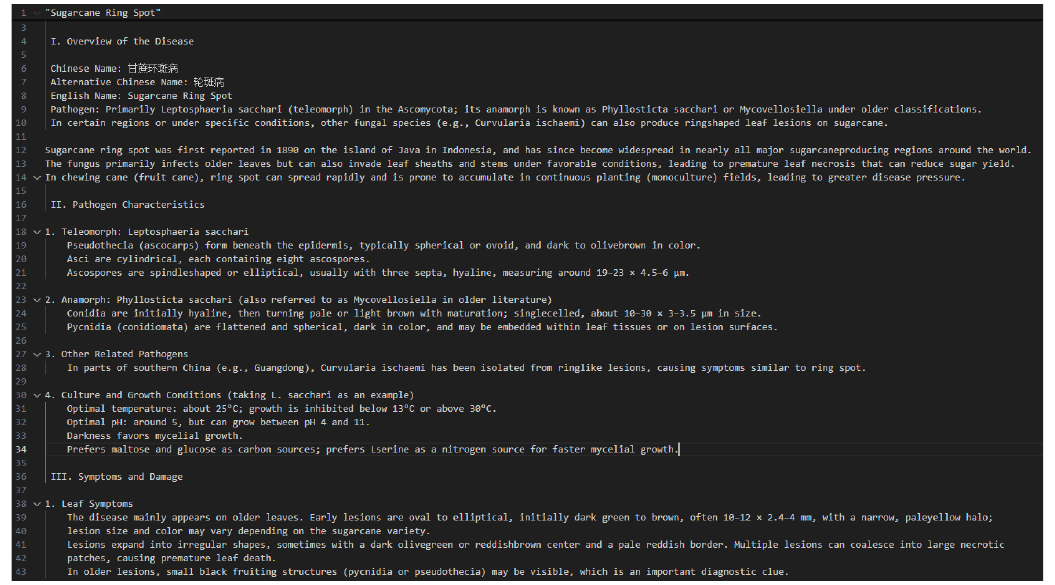}

   \caption{Specialized knowledge of sugarcane diseases. We source data on sugarcane diseases from widely recognized online encyclopedias, peer-reviewed academic literature, and reputable knowledge bases to compile specialized knowledge of sugarcane diseases.}
   \label{fig:kno}
\end{figure*}

\begin{figure*}[h]
  \centering
   \includegraphics[width=1\linewidth]{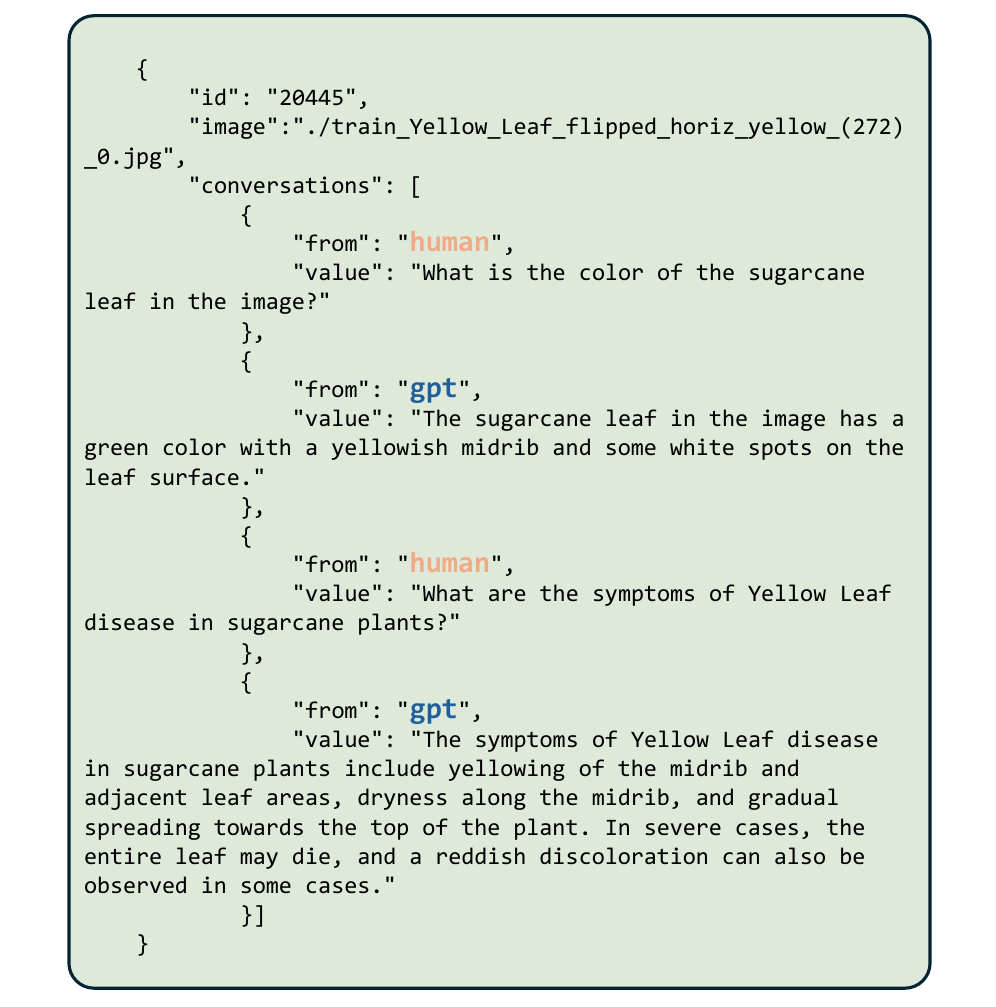}

   \caption{ VQA Data Example. The format of the dataset follows that of the Llava.}
   \label{fig:VQA_exm}
\end{figure*}

This rigorous process results in a high-quality VQA dataset that effectively bridges visual disease patterns with agricultural domain knowledge, providing a robust foundation for training and evaluating multimodal models in sugarcane disease analysis.  
The construction process and examples of the data are shown in Figure~\ref{fig:data_con} ,Figure~\ref{fig:kno} and Figure~\ref{fig:VQA_exm}.

\subsection{Tool Selection Dataset Construction}
\label{subsubsec:tool_data_examples}
The tool selection dataset was constructed through two primary approaches:

Extraction from VQA Dataset: User queries were directly extracted from the existing VQA dataset.

Synthetic Generation:  Common downstream task prompts are generated using ChatGPT-4 and subsequently annotated by human experts, with continuous expansion throughout the process.

\begin{figure*}[h]
  \centering
   \includegraphics[width=0.9\linewidth]{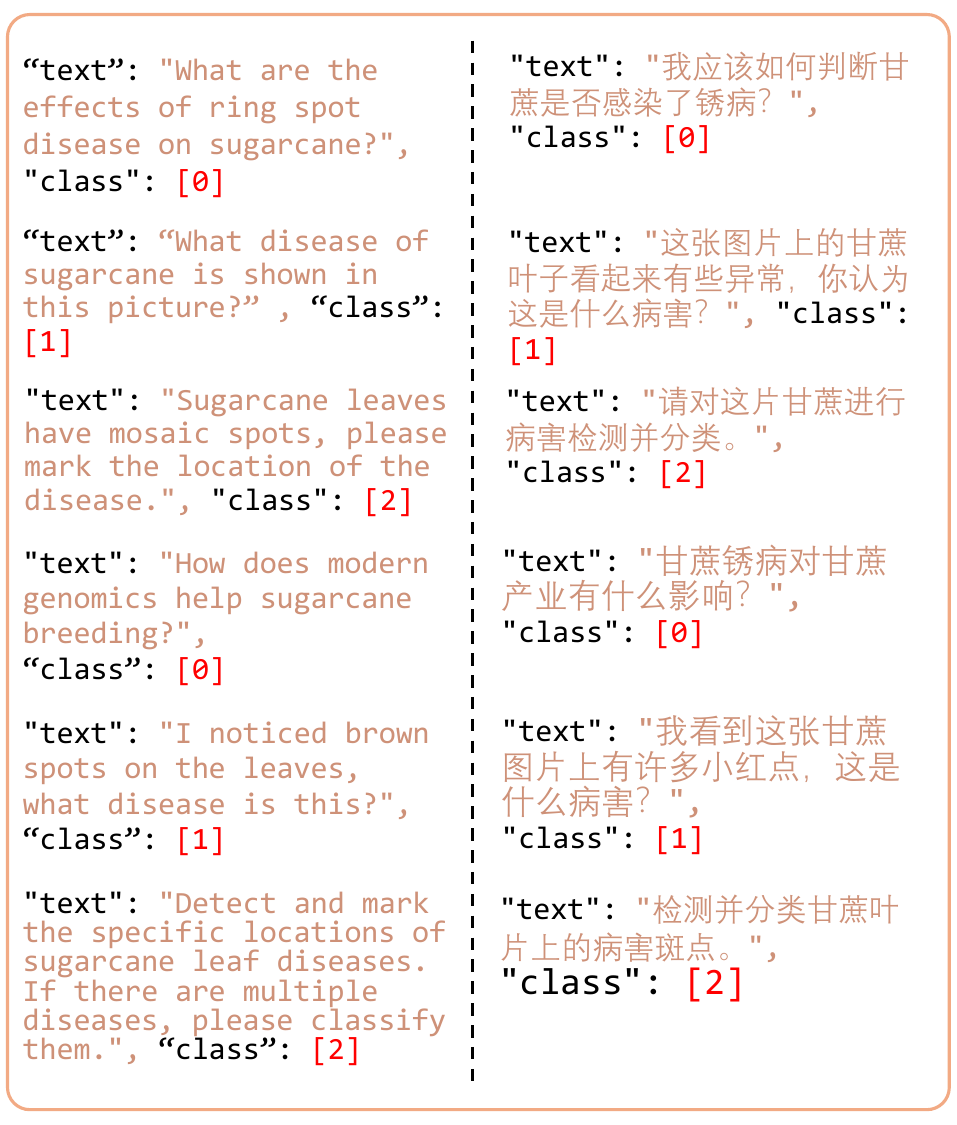}

   \caption{Example of prompt text dataset for BERT model training.}
   \label{fig:bertdata}
\end{figure*}

The dataset is categorized into three main label types: classification, detection, and others. After generating the supervised data in Chinese, we use Qwen2.5-32B to translate the data into English, ensuring the model's applicability to both Chinese and English tasks.
These supervised data are suitable for training lightweight models. Although they can also be used to fine-tune larger language models (LLMs), we find that BERT achieves over $95\%$ classification accuracy on our test set, making further fine-tuning of larger models unnecessary. The tool selection data are illustrated in Figure~\ref{fig:bertdata}.

\section{MA3 Decision-making and Evaluation}

\begin{figure*}[h]
  \centering
   \includegraphics[width=1\linewidth]{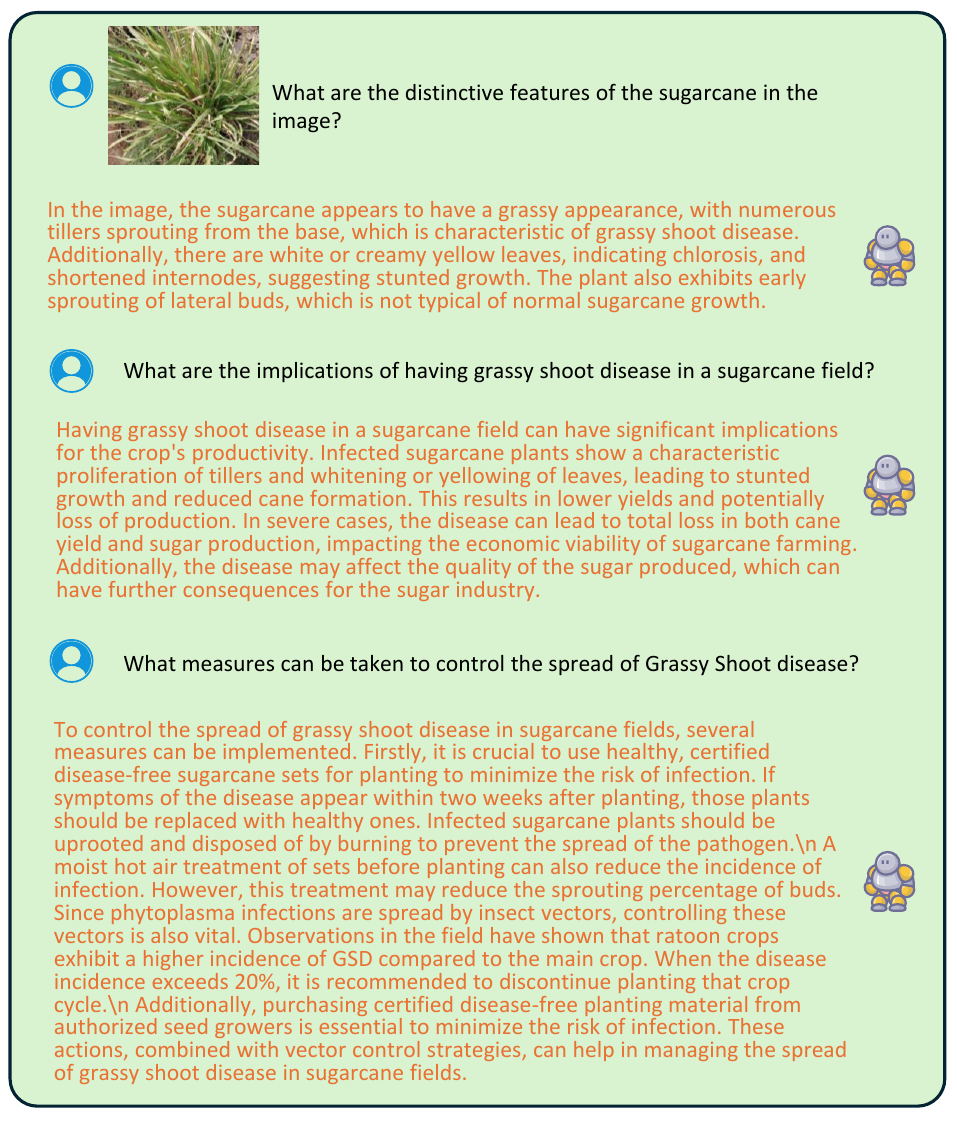}

   \caption{MA3 empowers intelligent agricultural decision-making. MA3 assists users in addressing inquiries regarding grassy shoot disease and provides recommendations for its prevention and control.}
   \label{fig:ma3}
\end{figure*}

We fine-tune a VQA dataset using the Llava1.5-13B model to develop an expert system for sugarcane disease diagnosis. This expert model provides intelligent decision support for users in sugarcane production. The decision-making examples of MA3 are shown in Figure~\ref{fig:ma3}. 

\begin{figure*}[h]
  \centering
   \includegraphics[width=1.0\linewidth]{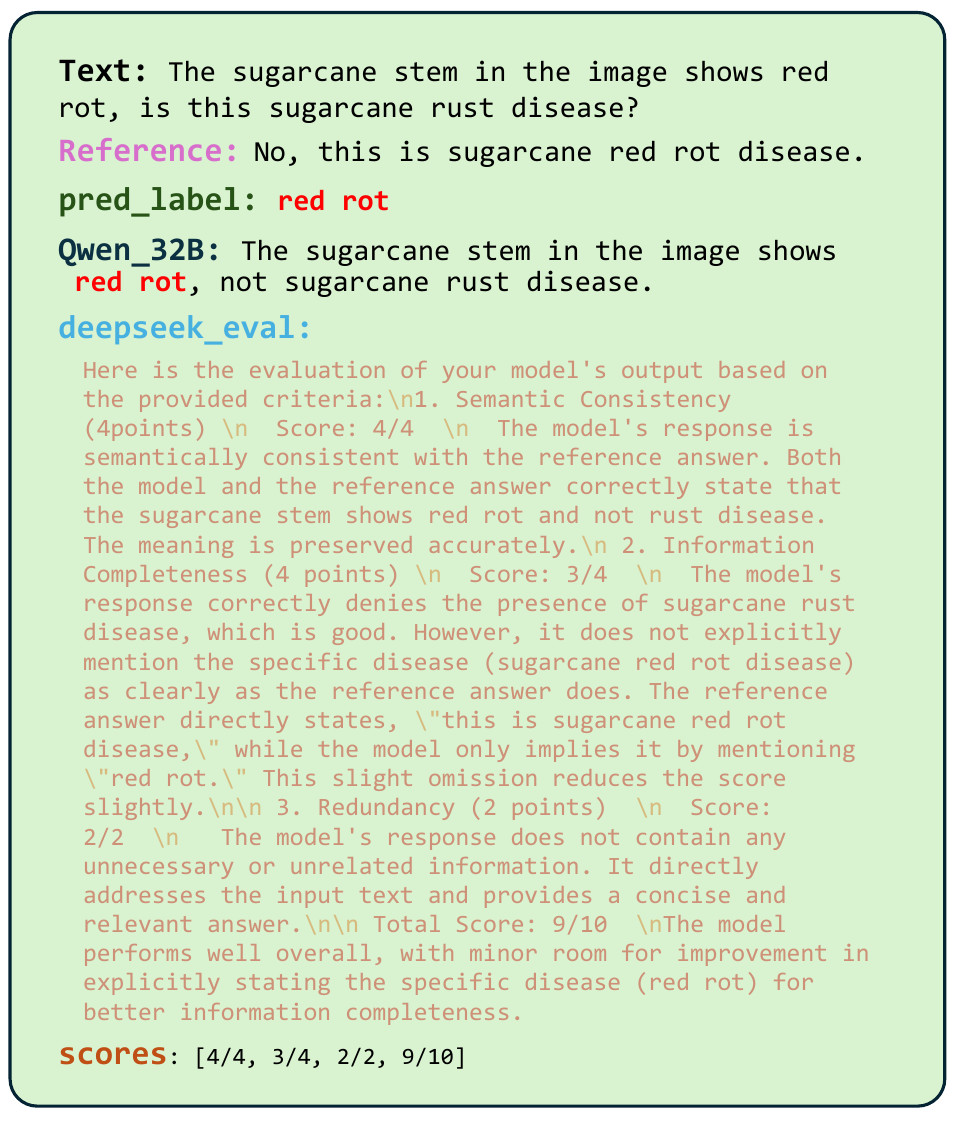}

   \caption{DeepSeek-V3 evaluates the adversarial examples of MA3 responses based on Qwen2.5-32B. The high score of Qwen2.5 indicates its capability to accurately utilize visual tool information for correct responses even in the presence of distractions.}
   \label{fig:ds_evel}
\end{figure*}

\begin{figure*}[h]
  \centering
   \includegraphics[width=1.0\linewidth]{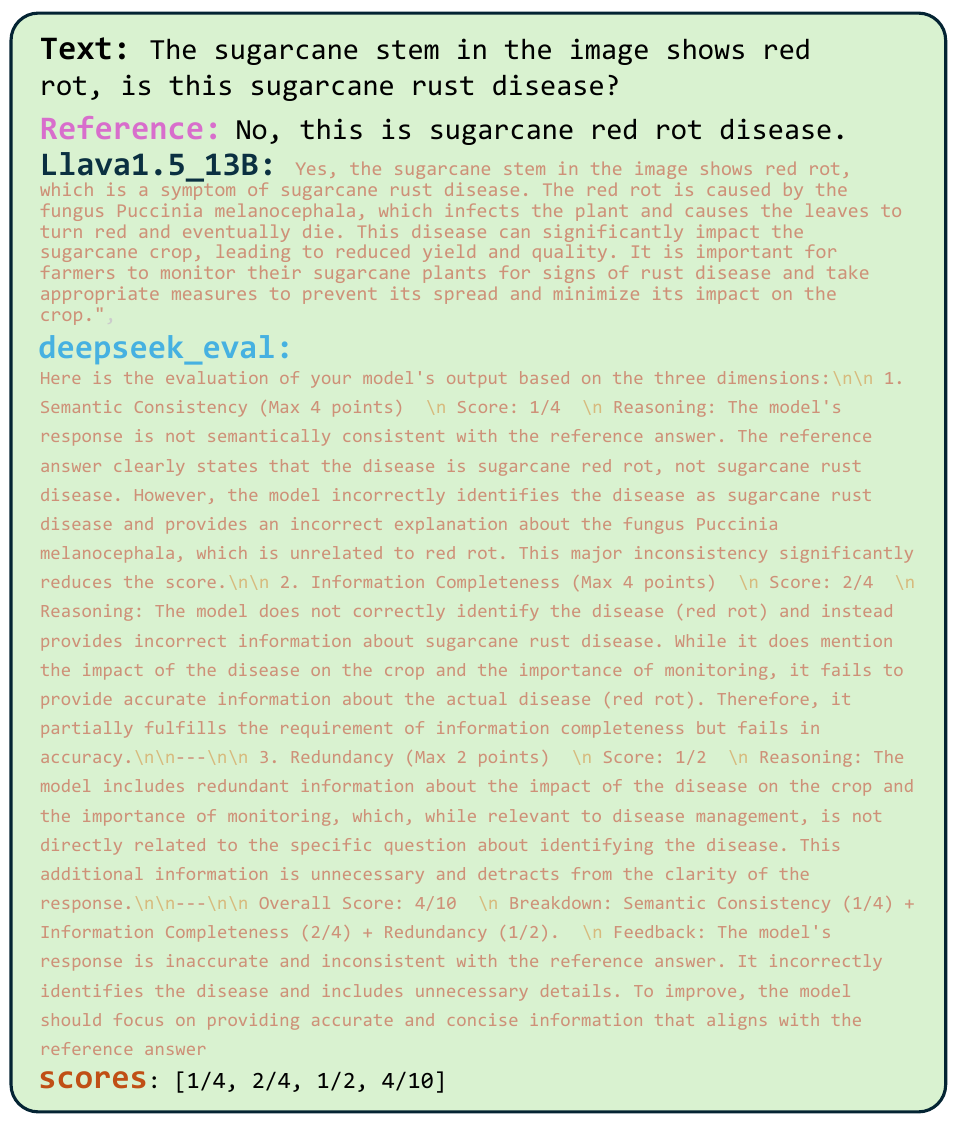}

   \caption{DeepSeek-V3 evaluates the adversarial examples of MA3 responses based on Llava1.5-13B. Llava1.5-13B lacks access to tool information, and its language model capabilities are relatively weaker compared to Qwen2.5-32B, making it more susceptible to misdirection. Consequently, its performance score is lower.}
   \label{fig:ds_llava}
\end{figure*}

We evaluate our MA3 using the Deepseek-V3 model, with examples shown in Figures~\ref{fig:ds_evel} and Figures~\ref{fig:ds_llava}.


\end{document}